%% file: icml_2024.tex
\documentclass{article}

\usepackage{microtype}
\usepackage{graphicx}
\usepackage{subfigure}
\usepackage{booktabs} % for professional tables
\usepackage{graphicx}
\usepackage{comment}
\usepackage{caption}
\usepackage{float}

\usepackage{graphicx} % Include this line in the preamble
\newif\ifcomments

\usepackage{hyperref}

\usepackage[accepted]{icml2024}

\usepackage{amsmath}
\usepackage{amssymb}
\usepackage{mathtools}
\usepackage{amsthm}
\usepackage{multicol}

\usepackage[capitalize,noabbrev]{cleveref}

\theoremstyle{plain}

\theoremstyle{definition}

\theoremstyle{remark}

\usepackage[textsize=tiny]{todonotes}

\input{math_commands.tex}

\input{macros}
\icmltitlerunning{HyperFields: Towards Zero-Shot Generation of NeRFs from Text}

\begin{document}

\twocolumn[
\icmltitle{HyperFields: Towards Zero-Shot Generation of NeRFs from Text}



\icmlsetsymbol{equal}{*}

\begin{icmlauthorlist}
\icmlauthor{Sudarshan Babu}{equal,ttic}
\icmlauthor{Richard Liu}{equal,uchicago}
\icmlauthor{Avery Zhou}{equal,ttic,uchicago}
\icmlauthor{Michael Maire}{uchicago}
\icmlauthor{Greg Shakhnarovich}{ttic}
\icmlauthor{Rana Hanocka}{uchicago}
\end{icmlauthorlist}

\icmlaffiliation{ttic}{Toyota Technological Institute at Chicago}
\icmlaffiliation{uchicago}{University of Chicago}

\icmlcorrespondingauthor{Sudarshan Babu}{sudarshan@ttic.edu}

\icmlkeywords{Machine Learning, ICML}

%\vskip 0.3in
\input{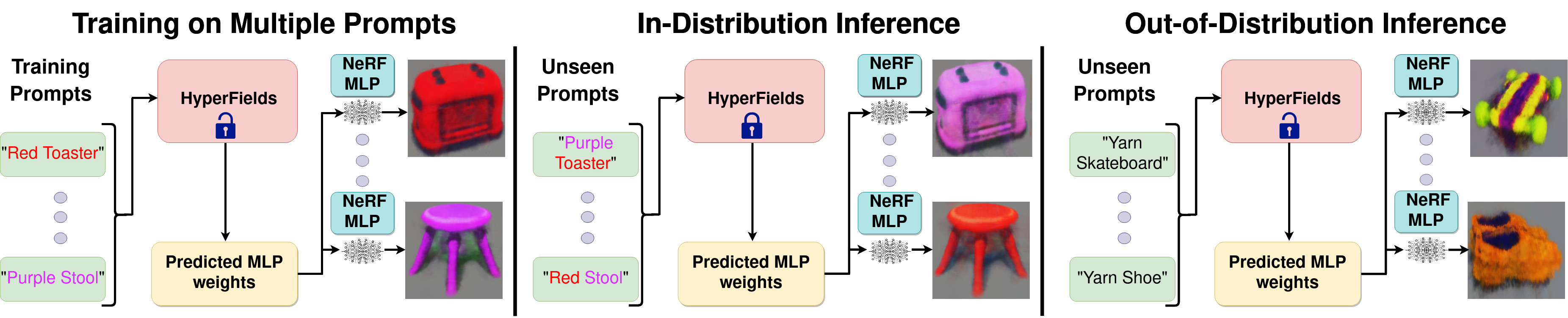}
]


\printAffiliationsAndNotice{\icmlEqualContribution} % otherwise use the standard text.

\input{abstract}
\input{introduction}

\input{related_work}
\input{method}
\input{results}
\input{conclusion}

\bibliography{iclr2024_conference}
\bibliographystyle{icml2024}
\newpage 
\input{supp_paper}
\end{document}

%% file: math_commands.tex
\usepackage{amsmath,amsfonts,bm}

\def\eqref#1{equation~\ref{#1}}
\def\1{\bm{1}}

\DeclareMathAlphabet{\mathsfit}{\encodingdefault}{\sfdefault}{m}{sl}
\SetMathAlphabet{\mathsfit}{bold}{\encodingdefault}{\sfdefault}{bx}{n}

%% file: macros.tex
\newcommand{\ourmethod}{HyperFields}

%% file: images/teaser.tex
\renewcommand\twocolumn[1][]{#1}%
\begin{center}
    \centering
	\includegraphics[width=0.95\textwidth]{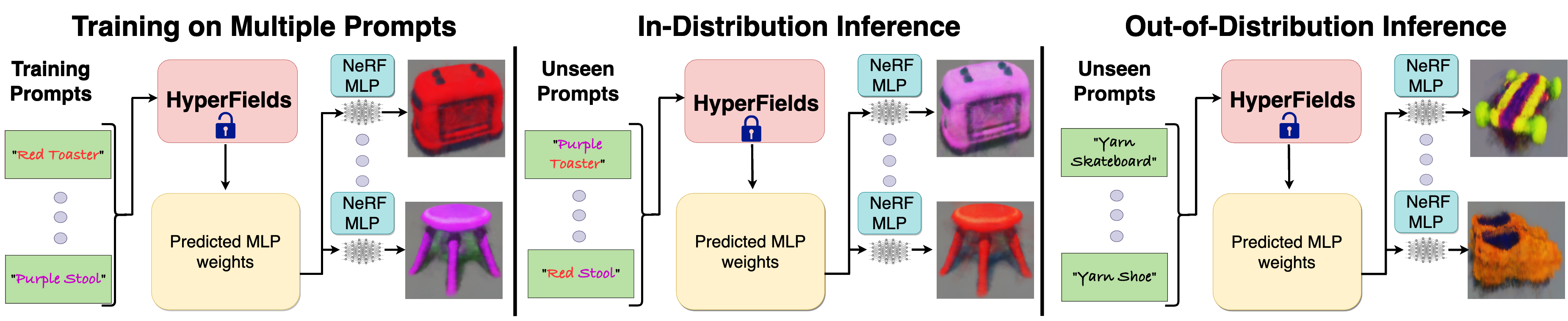}
	\vspace{-2mm}
	\captionof{figure}{\ourmethod{} is a hypernetwork that learns to map text to the space of weights of Neural Radiance Fields (first column). After training, HyperFields is capable of generating in-distribution scenes - \textit{unseen during training} - in a feed forward manner (second column), and for out-of-distribution prompts HyperFields can be fine-tuned to yield scenes respecting prompt semantics with just a few gradient steps (third column).}. 
    \label{fig:teaser}
    \vspace{2mm}%
\end{center}%

%% file: abstract.tex
\begin{abstract}
We introduce HyperFields, a method for generating text-conditioned Neural Radiance Fields (NeRFs) with a single forward pass and (optionally) some fine-tuning. Key to our approach are: (i) a dynamic hypernetwork, which learns a smooth mapping from text token embeddings to the space of NeRFs; (ii) NeRF distillation training, which distills scenes encoded in individual NeRFs into one dynamic hypernetwork. These techniques enable a single network to fit over a hundred unique scenes. We further demonstrate that HyperFields learns a more general map between text and NeRFs, and consequently is capable of predicting novel in-distribution and out-of-distribution scenes --- either zero-shot or with a few finetuning steps. Finetuning HyperFields benefits from accelerated convergence thanks to the learned general map, and is capable of synthesizing novel scenes 5 to 10 times faster than existing neural optimization-based methods. Our ablation experiments show that both the dynamic architecture and NeRF distillation are critical to the expressivity of HyperFields.
\end{abstract}

%% file: introduction.tex
\section{Introduction}
\label{sec:intro}

Recent advancements in text-to-image synthesis methods, highlighted by the works of \citet{dalle2, parti}, have ignited similar interest in the field of text-to-3D synthesis. This interest has grown in tandem with the emergence of Neural Radiance Fields (NeRFs) \citep{NeRF, pixelnerf, dietnerf}, which is a popular 3D representation for this task, due to their ability to robustly depict complex 3D scenes. 

To date, most text-conditioned 3D synthesis methods rely on either text-image latent similarity matching or diffusion denoising, both of which involve computationally intensive per-prompt NeRF optimization \citep{jain2021dreamfields, dreamfusion,lin2022magic3d}. Bypassing the need for per-prompt optimization remains a non-trivial challenge.

We propose to solve this problem through a hypernetwork-based neural pipeline, in which a single hypernetwork \citep{hypernetworks} is trained to generate the weights of individual NeRF networks, each corresponding to a unique scene. Once trained, this hypernetwork is capable of efficiently producing the weights of NeRFs corresponding to novel prompts, either through a single forward pass or with minimal fine-tuning. Sharing the hypernetwork across multiple training scenes enables effective transfer of knowledge to new scenes, leading to better generalization and faster convergence. However, we find that a naive hypernetwork design is hard to train. 

Our method, \emph{\ourmethod{}}, overcomes these challenges through several design choices. We propose predicting the weights of each layer of the NeRF network in a \emph{progressive} and \emph{dynamic} manner. Specifically, we observe that the intermediate (network) activations from the hypernetwork-predicted NeRF can be leveraged to guide the prediction of subsequent NeRF weights effectively. 

\begin{figure*}[t]
\begin{center}
  \includegraphics[width=\textwidth]{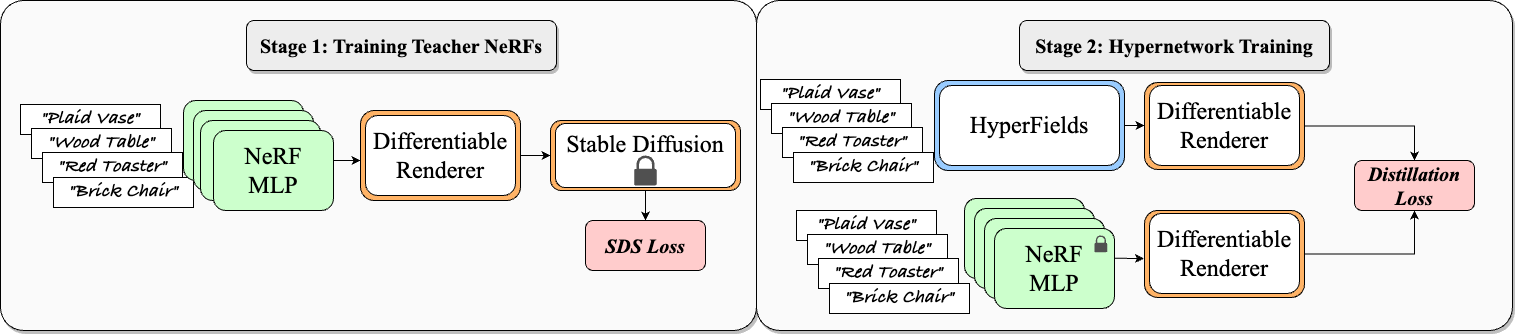}
  \caption{\textbf{Overview.} Our training pipeline proceeds in two stages. \textbf{Stage 1:} We train a set of single prompt text-conditioned teacher NeRFs using Score Distillation Sampling. \textbf{Stage 2:} We distill these single scene teacher NeRFs into the hypernetwork, through a photometric loss between the renders of the hypernetwork with the teacher network, which we dub our \textit{distillation loss.} \vspace{-0.5cm}}
  \label{fig:overview}
  \end{center}
\end{figure*}

To enhance the training of our hypernetwork, we introduce a distillation-based framework rather than the Score Distillation Sampling (SDS) used in \citet{dreamfusion,sjc}. We introduce NeRF distillation, in which we first train individual text-conditioned NeRF scenes (using SDS loss) that are used as teacher NeRFs to provide fine-grained supervision to our hypernetwork (see Fig.~\ref{fig:overview}). The teacher NeRFs provide exact colour and geometry labels, eliminating noisy training signals. 

% \sub{In addition to providing stability to training our hypernetwork, NeRF distillation also enables our model's renders to easily acquire the high generation quality provided by the more recent text-to-3D works. This is done by simply using the more recent works as the teacher NeRFs. Experimentally we show that our generated renders inherit the high generation quality.
% }

Our NeRF distillation framework allows for training \ourmethod{} on a much larger set of scenes than with SDS, scaling up to 100 different scenes without any degradation in scene quality. Importantly, NeRF distillation is agnostic to the choice of text-to-3D model, so that HyperFields can learn high-quality and complex scenes from the latest generative model in a plug-and-play fashion. We show results of our method trained on high-detail scenes from Prolific Dreamer in \cref{fig:complexpack,fig:att3d_ood,fig:att3d_compare}. 

% A potential explanation is that SDS loss exhibits high variance in loss signals throughout different sampling steps. This instability in the loss is likely a major challenge of training the hypernetwork on multiple scenes.

Once trained, our model can synthesize novel in-distribution NeRF scenes in a single forward pass (Fig.~\ref{fig:teaser}, second column) and enables accelerated convergence for out-of-distribution scenes, requiring only a few fine-tuning steps (Fig.~\ref{fig:teaser}, third column).  We clarify our use of the terms ``in-distribution" and ``out-of-distribution" in Sections \ref{sec:generalization} and \ref{sec:gen_ood} respectively. These results suggest that our method learns a semantically meaningful mapping. We justify our design choices through ablation experiments which show that both the dynamic hypernetwork architecture and NeRF distillation are critical to our model's expressivity.

Our successful application of dynamic hypernetworks to this difficult problem of generalized text-conditioned NeRF synthesis suggests a promising direction for future work on generalizing and parameterizing neural implicit functions through other neural networks. 
% \sub{
% Our contribuitions are two fold: 
% \begin{itemize}
%     \item \textbf{Dynamic HyperNetwork:} a novel hypernetwork design that provides flexibility required to learn a mapping between the space of text to space of NeRFs.
%     \item \textbf{NeRF Distillation:} a procedure to distill text-to-3D NeRFs into a single hypernetwork model, thereby providing stability to hypernetwork training and enabling the acquisition of high render quality exhibited by more recent text-to-3D works.
% \end{itemize}}

%% file: related_work.tex
\section{Background and Related Work}
Our work combines several prominent lines of work: neural radiance fields, score-based 3D synthesis, and learning function spaces using hypernetworks. 

% Our work can be seen as the combination of an implicit differentiable 3D representation, a frozen diffusion model, and an activation-conditioned dynamic hypernetwork. We begin by briefly overviewing existing methods to provide background prerequisites and a sense of the current literature.

\subsection{3D Representation via Neural Radiance Fields}
There are many competing methods of representing 3D data  in 3D generative modeling, such as point-clouds \citep{pointe, PointVoxelDiffusion}, meshes \citep{text2mesh, AvatarCLIP, LatentNeRF, LION}, voxels \citep{CLIPForge, TextCraft}, and signed-distance fields \citep{Neus, volsdf, kiloneus}. This work explores the popular representation of 3D scenes by Neural Radiance Fields (NeRF) \citep{NeRF, neuralfields, NeRF3DVision}. NeRFs were originally introduced to handle the task of multi-view reconstruction, but have since been applied in a plethora of 3D-based tasks, such as photo-editing, 3D surface extraction, and large/city-scale
3D representation~\citep{NeRF3DVision}.

There have been many improvements on the original NeRF paper, especially concerning training speed and fidelity \citep{PureCLIPNeRF, TensoRF, instantngp, directvoxel, plenoctree}. \ourmethod{} uses the multi-resolution hash grid introduced in InstantNGP~\citep{instantngp}.

% Last, as inspired by previous works that experimented with packing multiple scenes into the same NeRF \cite{pixelnerf, park202nerfies, park2021hypernerf}, our work treats the MLP as not merely representing a single scene, but rather as an indexable representation of many scenes. However, rather than indexing the MLP with some form of scene id, we use the text embedding $\epsilon (T)$ to query for the desired scene. 

\subsection{Score-Based 3D Generation}
While many works attempt to directly learn the distribution of 3D models via 3D data, others opt to use guidance from 2D images due to the vast difference in data availability. Such approaches replace the photometric loss in NeRF's original objective with a guidance loss. The most common forms of guidance in the literature are from CLIP \citep{clip} or a frozen, text-conditioned 2D diffusion model. The former methods minimize the cosine distance between the image embeddings of the NeRF's renderings and the text embedding of the user-provided text prompt \citep{jain2021dreamfields, PureCLIPNeRF, dietnerf}.

Noteworthy 2D diffusion-guided models include DreamFusion \citep{dreamfusion} and Score Jacobian Chaining (SJC) \citep{sjc}, which feed noised versions of images rendered from a predicted NeRF into a frozen text-to-image diffusion model (Imagen \citep{imagen} and StableDiffusion \citet{rombach2021highresolution}, respectively) to obtain what can be understood as a scaled Stein Score \citep{stein}. Our work falls into this camp, as we rely on score-based gradients derived from StableDiffusion to train the NeRF models which guide our hypernetwork training.

We use the following gradient motivated in DreamFusion:
\begin{equation} 
\nabla_{\theta} \mathcal{L}(\phi,g(\theta)) \triangleq  \mathbb E_{t,c} 
\begin{bmatrix} 
	w(t) (\hat \epsilon_\phi(z_t; y, t)-\epsilon)\frac{\partial x}{\partial \theta})
\end{bmatrix}
\end{equation}
which is similar to the gradient introduced in SJC, with the key difference being SJC directly predicts the noise score whereas DreamFusion predicts its residuals. We refer to optimization using this gradient as \textit{Score Distillation Sampling} (SDS), following the DreamFusion authors. Followup work has aimed at improving 3D generation quality \citep{wang2023prolificdreamer, metzer2023latent,chen2023fantasia3d}, whereas we target an orthogonal problem of generalization and convergence of text-to-3D models.

%\rl{I'd recommend putting this to the end of the Introduction section for visibility. I would also say we should stick to the ``concurrent and independent" claim while highlighting that we can do accelerated out-of-distribution convergence, which indicates we learn a general semantic map. I wouldn't mention the proprietary model/lack of code and reproducibility unless the reviewers bring it up}. 
\textbf{Connections to ATT3D:}
We note that our work is concurrent and independent of ATT3D \citep{lorraine2023att3d}. We are similar in that we both train a hypernetwork to generate NeRF weights for a set of scenes during training and generalize to novel in-distribution scenes without any test time optimization.  On top of the in-distribution generalization experiments, we also demonstrate accelerated convergence to novel out-of-distribution scenes (defined in \ref{sec:gen_ood}), which ATT3D does not.

On the technical side, we primarily differ in our novel dynamic hypernetwork architecture. Our hypernetwork generates the MLP weights of the NeRF, while ATT3D outputs the weights of the hash grid in their InstantNGP model. Importantly, our hypernetwork layers are conditioned on not just the input text prompt, but also the activations of the generated NeRF MLP (\ref{sec:dynamic}). We show through our ablations that this dynamic hypernetwork conditioning is essential to the expressivity of our network, as it enables the network to change its weights for the same scene as a function of the view that is being rendered. In contrast, in ATT3D, the generated hash grid is the same regardless of the view being rendered, potentially resulting in the loss of scene detail.

Finally, ATT3D is built on Magic3D \citep{lin2022magic3d} which is a proprietary and more powerful text-to-3D model than the publicly available stable DreamFusion model \citep{stable-dreamfusion} that we use in most of our experiments. We show that our model is capable of learning high quality and complex NeRF scenes produced by more powerful models such as ProlificDreamer without reduction in generation quality \ref{sec:prolificdreamer}.

\subsection{HyperNetworks}
Hypernetworks are networks that are used to generate weights of other networks which perform the actual task (task performing network) \citep{ha2016hypernetworks}. Many works attempt to use hypernetworks as a means to improve upon conditioning techniques. Among these, some works have explored applying hypernetworks to implicit 2D representations \citep{siren, FiLM, hyperstyle}, and 3D representations \citep{scenerepresentationnetworks, lightfield, StylizingNeRF}. Very few works apply hypernetworks to radiance field generation. Two notable ones are HyperDiffusion and Shape-E, which both rely on denoising diffusion for generation \citep{erkoc_hyperdiffusion_2023, jun_shap-e_2023}. HyperDiffusion trains an unconditional generative model which diffuses over sampled NeRF weights, and thus cannot do text-conditioned generation. Shap-E diffuses over latent codes which are then mapped to weights of a NeRF MLP, and requires teacher point clouds to train. Due to the memory burden of textured point clouds, scene detail is not well represented in Shap-E. Both of these methods have the same limitations of slow inference due to denoising sampling. In contrast, our method predicts NeRF weights dynamically conditioned on the 1) text prompt, 2) the sampled 3D coordinates, and 3) the previous NeRF activations.

An interesting class of hypernetworks involve models conditioned on the activations or inputs of the task-performing network \citep{chen2020dynamic}. These models take the following form: let $h,g$ be the hypernetwork and the task performing network respectively. Then $W = h(a)$, where $W$ acts as the weights of $g$ and $a$ is the activation from the previous layer of $g$ or the input to $g$. These are called dynamic hypernetworks, as the predicted weights change dynamically with respect to the layer-wise signals in $g$. Our work explores the application of dynamic hypernetworks to learning a general map between text and NeRFs.

%% file: method.tex
\section{Method}
\label{sec:dynamic}
Our method consists of two key innovations, the dynamic hypernetwork architecture and NeRF distillation training. We discuss each of these two components in detail below.

\begin{figure*}[t]
    \centering
    \includegraphics[width=0.8\textwidth]{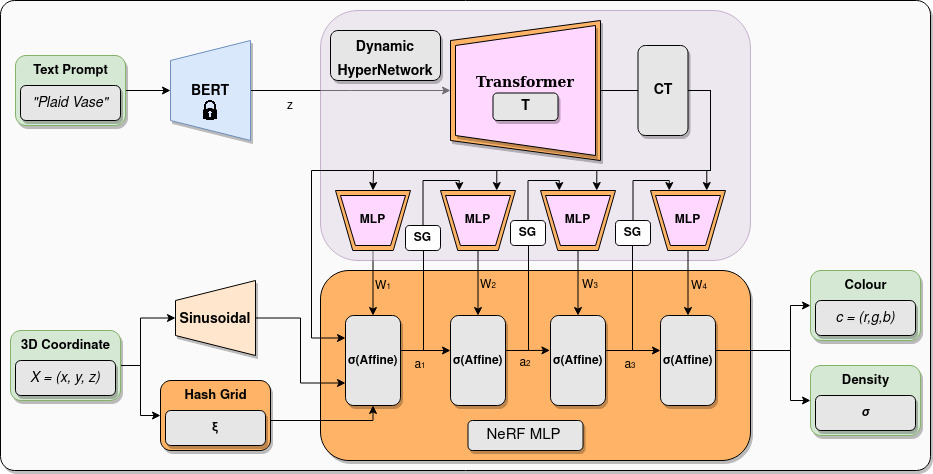}
    \caption{The input to the \ourmethod{} system is a text prompt, which is encoded by a pre-trained text encoder (frozen BERT model). The text latents are passed to a Transformer module, which outputs a conditioning token (CT). This conditioning token (which supplies scene information) is used to condition each of the MLP modules in the hypernetwork. The first hypernetwork MLP (on the left) predicts the weights $W_1$ of the first layer of the NeRF MLP. The second hypernetwork MLP then takes as input both the CT and $a_1$, which are the \textit{activations} from the first predicted NeRF MLP layer, and predicts the weights $W_2$ of the second layer of the NeRF MLP. The subsequent scene-conditioned hypernetwork MLPs follow the same pattern, taking the activations $a_{i-1}$ from the previous predicted NeRF MLP layer as input to generate weights $W_{i}$ for the $i^{th}$ layer of the NeRF MLP. We include stop gradients (SG) to stabilize training.\vspace{-0.2cm}}
  \label{fig:high_level}
\end{figure*}

\subsection{Dynamic Hypernetwork}
%\label{sec:dynamic}
The dynamic hypernetwork consists of the Transformer $\mathcal{T}$ and MLP modules as given in figure \ref{fig:high_level}. The sole input to the dynamic hypernetwork is the scene information represented as a text description. The text is then encoded by a frozen pretrained BERT model, and the text embedding $z$ is processed by $\mathcal{T}$. Let conditioning token CT = $\mathcal{T}(z)$ be the intermediate representation used to provide the current scene information to the MLP modules. Note that the text embeddings $z$ can come from any text encoder, though in our experiments we use frozen BERT embeddings.

In addition to conditioning token CT, each MLP module takes in the activations from the previous layer $a_{i-1}$ as input.  Given these two inputs, the MLP module is tasked with generating parameters $W_{i}$ for the $i^{th}$ layer of the NeRF MLP. For simplicity let us assume that we sample only one 3D coordinate and viewing direction per minibatch, and let $h$ be the hidden dimension of the NeRF MLP. Then $a_{i-1} \in \mathbb{R}^{1 \times h}$.  Now the weights $W_{i} \in \mathbb{R}^{h \times h}$  of the $i^{th}$ layer are given as follows: %\vspace{-0.3cm}
\begin{eqnarray}
W_{i} = \text{MLP}_{i}(CT,a_{i-1}) 
\end{eqnarray}
\vspace{-0.1cm}%
The forward pass of the $i^{th}$ layer is:
\begin{eqnarray}
%\overline{a}_{i-1} = \mu(a_{i-1}) \\
%W_{i} = \text{MLP}_{i}(e,\overline{a}_{i-1})
a_{i} = W_{i}*a_{i-1} \label{eqn:fp}
\end{eqnarray}
where $a_{i} \in \mathbb{R}^{1\times h}$ and * is matrix multiplication. This enables the hypernetwork MLPs to generate a different set of weights for the NeRF MLP that are best suited for each given input 3D point and viewing direction pair. This results in effectively a unique NeRF MLP for each 3D point and viewing direction pair.

In practice training with minibatch size 1 is impractical, so during training we sample a non-trivial minibatch size and generate weights that are best suited for the given minibatch, as opposed to weights unique to each 3D coordinate and viewing direction pair as illustrated above.

In order to generate a unique set of weights for a given minibatch we do the following:
\begin{eqnarray}
\overline{a}_{i-1} =& \mu(a_{i-1}) \\
W_{i} =& MLP_{i}(CT,\overline{a}_{i-1})
%W_{i} = MLP_{i}(e,a_{i-1}) 
\end{eqnarray}
Where $\mu(.)$ averages over the minibatch index. So if the minibatch size is $n$, then $a_{i-1} \in R^{n\times h}$, and $\overline{a}_{i-1} \in \mathbb{R}^{1\times h}$ and the forward pass is still computed as given in equation \ref{eqn:fp}. This adaptive nature of the predicted NeRF MLP weights leads to the increased flexibility of the model. As shown in our ablation experiments in \cref{abl:packing}, it is an essential piece to our model's large scene capacity. %\vspace{-0.2cm}
\subsection{NeRF Distillation}
%\vspace{-0.2cm}
As shown in figure \ref{fig:overview}, we first train individual DreamFusion NeRFs on a set of text prompts, following which we train the HyperFields architecture with supervision from these single-scene DreamFusion NeRFs. 

The training routine is outlined in Algorithm \ref{alg:training}, in which at each iteration, we sample $n$ prompts and a camera viewpoint for each of these text prompts (lines 2 to 4). Subsequently, for the $i^{th}$ prompt and camera viewpoint pair we render image $\mathcal{I}_{i}$ using the $i^{th}$ pre-trained teacher NeRF (line 5). We then condition the HyperFields network $\phi_{hf}$ with the $i^{th}$ prompt, and render the image $I^{'}_{i}$ from the $i^{th}$ camera view point (line 6). We use the image rendered by the pre-trained teacher NeRF as the ground truth supervision to HyperFields (line 7). For the same sampled $n$ prompts and camera viewpoint pairs, let $\mathcal{I}^{'}_{1}$ to $\mathcal{I}^{'}_{n}$ be the images rendered by HyperFields and $\mathcal{I}_{1}$ to $\mathcal{I}_{n}$ be the images rendered by the respective pre-trained teacher NeRFs. The distillation loss is given as follows:
\begin{eqnarray}
\mathcal{L}_{d} = \frac{\sum_{i=1}^{n} (I_{i} - I^{'}_{i})^{2}}{n}
\end{eqnarray}
We observe through our ablations in \cref{abl:distillation} that this simple distillation scheme greatly helps HyperFields in learning to fit multiple text prompts simultaneously, as well as learn a more general mapping of text to NeRFs.

%\subsection{Implementation Details}
%We use the multiresolution hash grid developed in InstantNGP \citet{instantngp} for its fast inference with low memory overhead, and sinusoidal encodings $\gamma$ to combat the known spectral bias of neural networks \citep{spectralbias}. The NeRF MLP has 6 layers (with weights predicted by the dynamic hypernetwork), with skip connections every two layers. The dynamic hypernetwork MLP modules are two-layer MLPs with ReLU non-linearities and the Transformer module has 6 self-attention layers. Furthermore, we perform adaptive instance normalization before passing the activations into the MLP modules of the dynamic hypernetwork and also put a stop gradient operator on the activations being passed into the MLP modules (as in figure \ref{fig:high_level}). The exact dimensions of the various components of the architecture are described in the supplemental material.

%% file: results.tex
\section{Results}
\begin{figure}[h]
    \centering
    \includegraphics[width=0.5\textwidth]{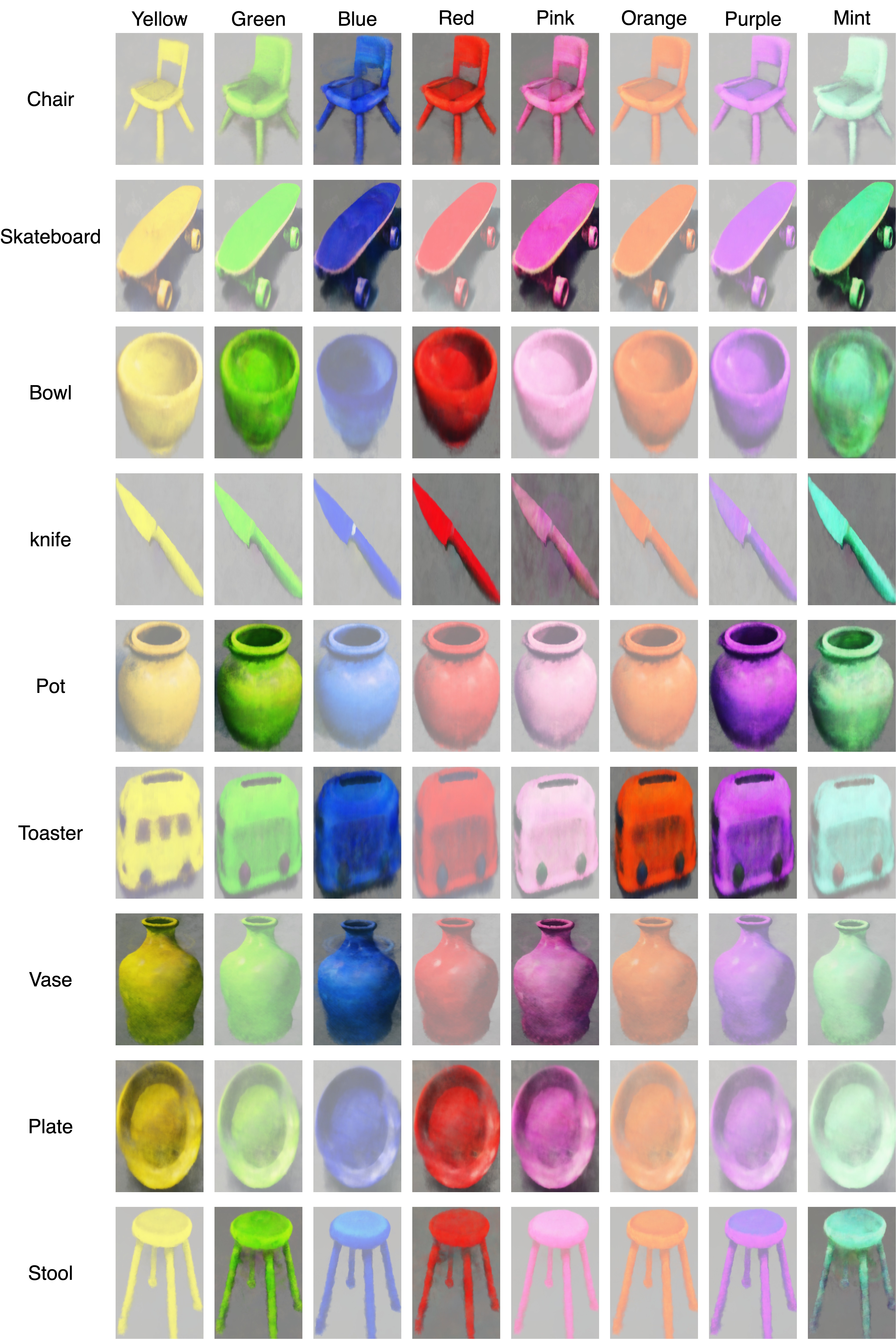}
    \caption{\textbf{Zero-Shot In-Distribution Generalization.} We train \ourmethod{} on a 9x8 grid of object/color combination scenes, and hold out a subset of combinations. The faded scenes are in the training set and the bright scenes are the trained model's \textbf{zero-shot predictions} of the holdout set.}
    % \textbf{Please take note that while we are showcasing a subset of our zero-shot generation results (3 objects across 8 different colors) here, our comprehensive demonstration includes zero-shot generation across a total of 9 objects spanning 8 colors. For a more detailed overview, we kindly direct your attention to Appendix Figure ~\ref{}.}}
    \label{fig:colormatrix}
\end{figure}

We evaluate \ourmethod{} by demonstrating its generalization capabilities, out-of-distribution convergence, amortization benefits, and ablation experiments. In Sec.~\ref{sec:generalization} and Sec.~\ref{sec:gen_ood} we evaluate the model's ability to synthesize novel scenes, both in and out-of-distribution. We quantify the amortization benefits of having this general model compared to optimizing individual NeRFs in Sec.~\ref{sec:amortization}. Finally, our ablations in Sec.~\ref{sec:ablations} justify our design choices of dynamic conditioning and NeRF distillation training.

\subsection{In-Distribution Generalization}
\label{sec:generalization}
%\subsubsection{In-Distribution: Seen Shape, Seen Color, Unseen Combination}
Fig.~\ref{fig:colormatrix} shows the results of training on a subset of combinations of 9 shapes and 8 colours, while holding out 3 colours for each shape. Our model generates NeRFs in a zero-shot manner for the held-out prompts (opaque scenes in Fig.~\ref{fig:colormatrix}) with quality nearly identical to the trained scenes. 

We call this \textit{in-distribution generalization} as both the shape and the color are seen during training but the inference scenes (opaque scenes in Fig.\ref{fig:colormatrix}) are novel because the combination of color and shape is unseen during training. For example: ``Orange toaster" is a prompt the model has not seen during training, though it has seen the color ``orange" and the shape ``toaster" in its training set. 
%We quantitatively evaluate the quality of our zero-shot predictions with CLIP retrieval scores. The support set for the retrieval consists of all 72 scenes (21 unseen and 35 seen) shown in Fig.~\ref{fig:colormatrix}. In Table ~\ref{tab:clip_ret} we compute the top-$k$ retrieval scores by CLIP similarity. The table reports the average scores for Top-1, 3, 5, 6, and 10 retrieval, separated by unseen (zero-shot) and seen prompts. The similarity in scores between the unseen and seen prompts demonstrates that our model's zero-shot predictions are of similar quality to the training scenes with respect to CLIP similarity.

We quantitatively evaluate the quality of our zero-shot predictions with CLIP retrieval scores. The support set for the retrieval consists of all 72 scenes (27 unseen and 45 seen) shown in Fig.~\ref{fig:colormatrix}. In Table~\ref{tab:clip_ret} we compute the top-$k$ retrieval scores by CLIP similarity. The table reports the average scores for top-$k$ retrieval, separated by unseen (zero-shot) and seen prompts. The similarity in scores between unseen and seen prompts indicates that our model's zero-shot predictions are of similar quality to the training scenes.

\begin{table}
  \begin{tabular}{lccccc}
    \toprule
    & Top-1 & Top-3 & Top-5 & Top-6 &  Top-10 \\
    \midrule
    Unseen & 57.1 & 85.7 & 85.7 & 90.4 & 95.2 \\
    Seen & 69.5 & 88.1 & 94.9 &94.9 &96.6 \\
    \bottomrule
  \end{tabular}
  \caption{CLIP Retrieval Scores: We report the average retrieval scores for the scenes shown in Fig.~\ref{fig:colormatrix}. We achieve similar scores between the seen and unseen prompts, indicating that our zero-shot generations are of similar quality to the training scenes.}
  \label{tab:clip_ret}
  \end{table}

%\begin{figure*}
%    \centering
    %\includegraphics[width=18cm]{images/compare_converge_clear_big_fixed.png}
    %\includegraphics[width=18cm]{images/main.png}
%    \includegraphics[width=18cm]{images/ood_compare_v2.jpg}
%    \caption{\textbf{Comparing geometries at convergence:} In this experiment we simply let the baselines (row 2 and 3) in Figure \cref{fig:ood} continue training till the scene generated is reasonably similar to the text prompt. However, despite training the baseline models longer, in quite a few cases the geometry/texture generated by our method is better than the baselines. We compare and discuss our quality advantage in \cref{sec:ood}. The inset figures top left hand side are the zero shot predictions by our model, and in the case of the fine tunes baseline it is the scene the NeRF is initially trained on.}
%    \label{fig:oodvisual}
%\end{figure*}

\begin{figure*}[ht!]
    \centering
    \includegraphics[width=\textwidth]{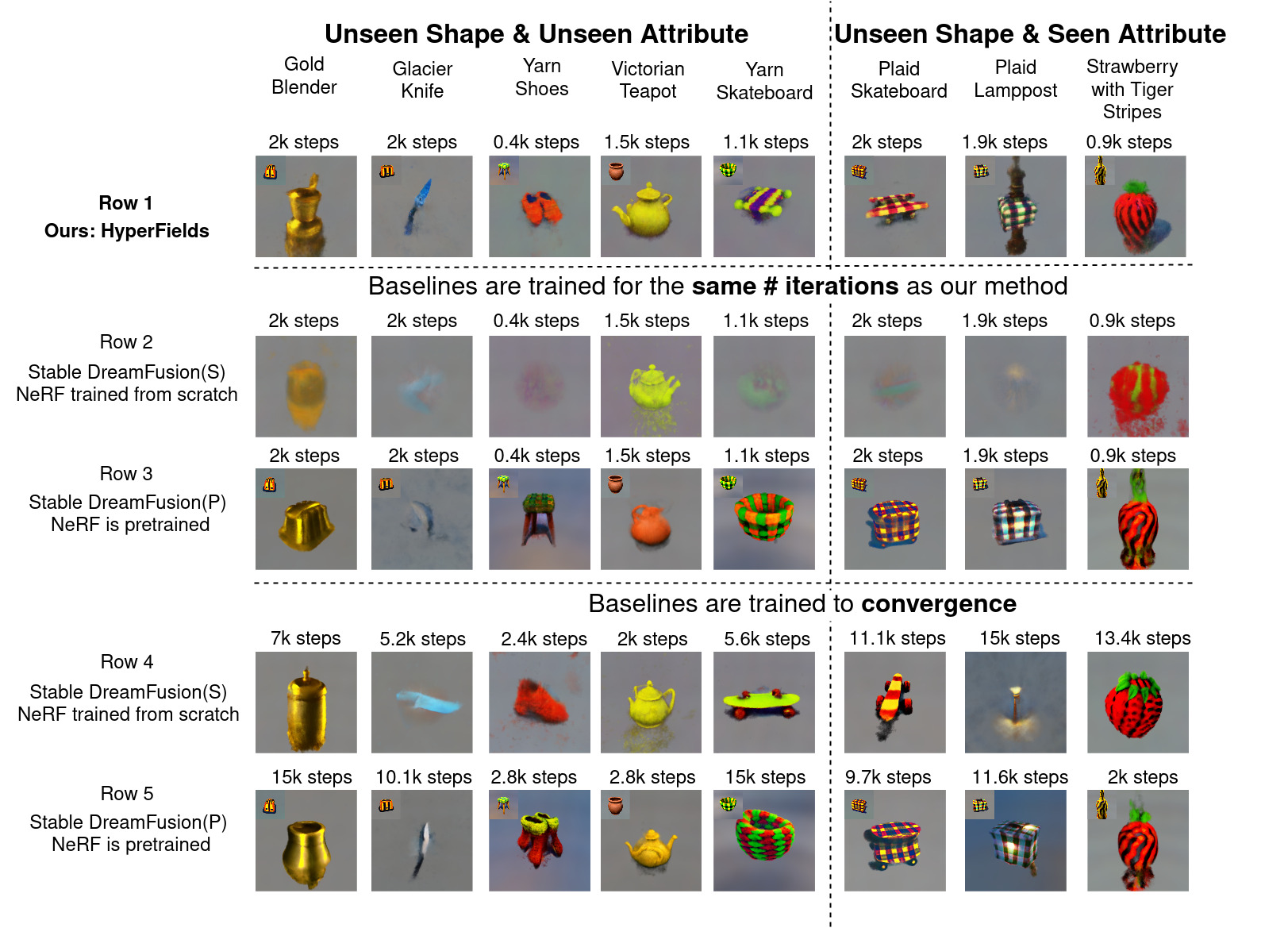}
    % \includesvg[width=20cm]{images/ood_compare_2.svg}
    \caption{\textbf{Finetuning  to out-of-distribution prompts: unseen shape and or unseen attribute. } Our method generates out-of-distribution scenes in at most 2k finetuning steps (row 1), whereas the baseline models are far from the desired scene at the same number of iterations (rows 2 and 3). When allowed to fine-tune for significantly longer (rows 4 and 5) the baseline generations are at best comparable to our model's generation quality, demonstrating that our model is able to adapt better to out-of-distribution scenes.\vspace{-0.1cm}
}
    \label{fig:ood}
\end{figure*}

\begin{table*}[h!]
    \centering
    \resizebox{\textwidth}{!}{
        \begin{tabular}{ccccccc}
        \toprule
        Model & \parbox{1cm}{\centering Golden\\ Blender}  & \parbox{1cm}{\centering Yarn\\ Shoes} & \parbox{1.5cm}{\centering Yarn \\Skateboard} & \parbox{1.5cm}{\centering Plaid \\Skateboard} & \parbox{1.5cm}{\centering Plaid\\ Lamppost} & \parbox{2cm}{\centering Strawberry \\with tiger stripes} \\
        \midrule
        Our Method ($\downarrow$) & $1.3 \pm 0.14$  & $1.0 \pm 0.09$ & $1.3 \pm 0.07$ & $1.3 \pm 0.11$ & $1.4 \pm 0.03$ & $1.1 \pm 0.14$ \\
        Best DreamFusion Baseline ($\downarrow$) & $2.5 \pm 0.11$ & $2.4 \pm 0.13$ & $2.3 \pm 0.11$ & $1.7 \pm 0.09$ & $2.0 \pm 0.09 $ & $2.2 \pm 0.08$ \\
        P-Score ($\downarrow$) & $1.0 \times 10^{-25}$ & $2.01 \times 10^{-34}$  & $2.8 \times 10^{-30}$ & $1.1 \times 10^{-8}$ & $1.3 \times 10^{-25}$ & $2.0 \times 10^{-26}$ \\
        \bottomrule
        \end{tabular}
    }
    \caption{\textbf{Average User-Reported Ranks (N=450):} We report the average rank submitted by all users for our method, and compute the average rank for all 33 of the baselines. We report the average rank of the best performing baseline for each prompt (with $\pm$95\% confidence intervals). Our method is consistently preferred over the best baseline, despite the best baseline consuming 33x more computational resources than our method to find. We report the p-value for the difference in rank between our method and the next best DreamFusion baseline, and find it is significant at the 1\% level across all our prompts.}
    \label{tab:study}
\end{table*}
\begin{figure*}[t]
\centering
\includegraphics[width=0.7\textwidth]{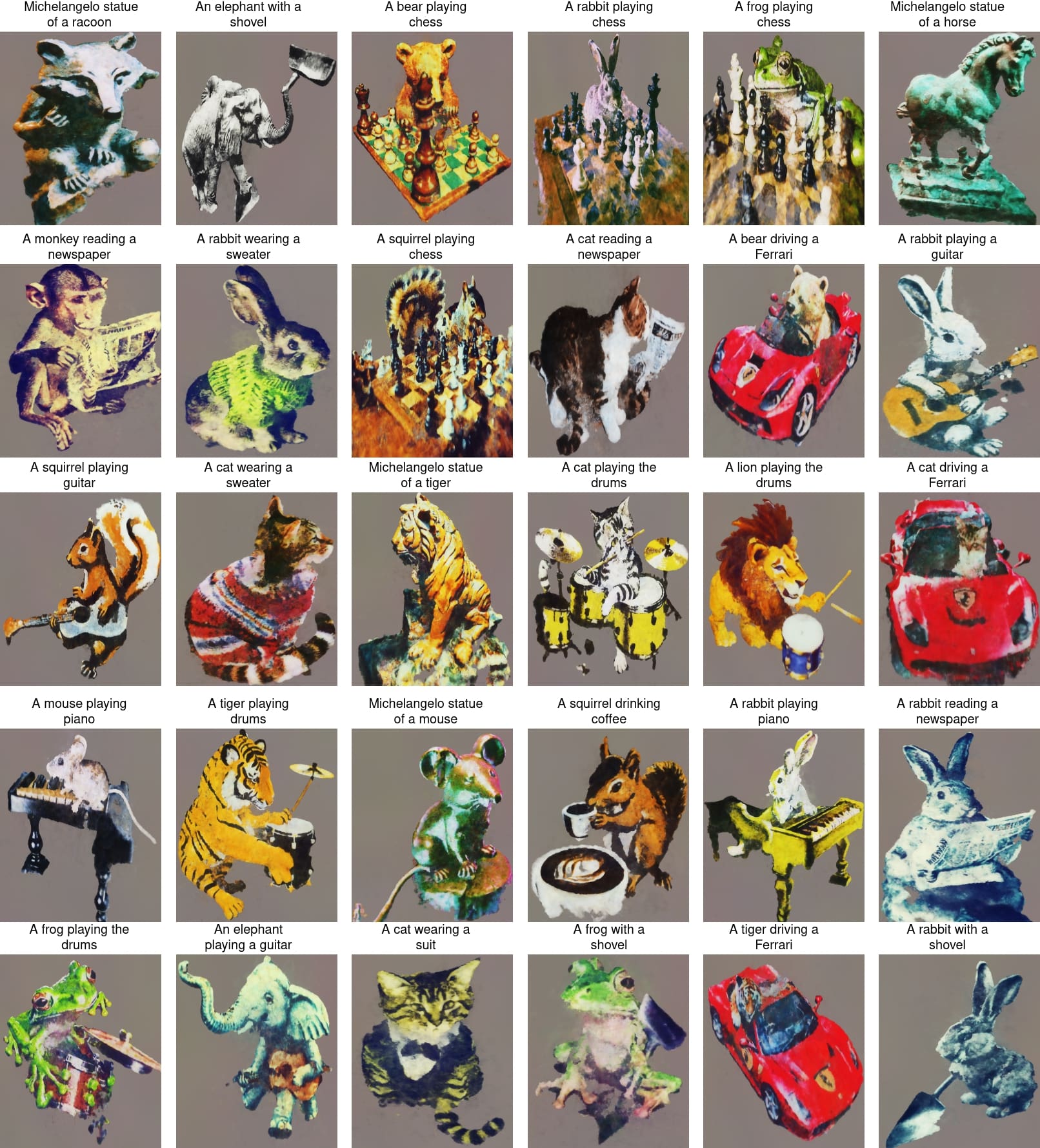}
\caption{\textbf{Prolific Dreamer scenes distilled into HyperFields:} We distill 30 high quality and complex scenes generated by Prolific Dreamer into a single HyperFields model, which underscores the modeling capacity of our novel architecture.} 
\label{fig:adddreamerfields}
\end{figure*}

\subsection{Accelerated Out-of-Distribution Convergence}
\label{sec:gen_ood}
We further test \ourmethod{}'s ability to generate shapes and attributes that it has \textit{not seen} during training. We call this \textit{out-of-distribution inference} because the specified geometry and/or attribute are not within the model's training set.

We train our model on a rich source of prompts, across multiple semantic dimensions (material, appearance, shape). The list of prompts used is provided in the appendix material section~\ref{sec:ood_supp} using NeRF distillation loss. Post training, we test our model on the prompts in Fig.~\ref{fig:ood}. The prompts are grouped based on whether both shape and attribute are unseen (column 1, Fig.~\ref{fig:ood}) or just the shape is unseen (column 2, Fig.~\ref{fig:ood}). For example, in ``gold blender" both material ``gold" and shape ``blender" are unseen during training. 

Since these prompts contain geometry/attributes that are unseen during training, we do not expect high quality generation without additional optimization. Instead, we demonstrate that fine-tuning the trained \ourmethod{} model on SDS loss for the given the out-of-distribution prompt can lead to accelerated convergence especially when compared to the DreamFusion baselines. 

We consider two baselines, 1) \textbf{Stable Dreamfusion (S):} Publicly  available implementation of Dreamfusion trained from \textbf{S}cratch, 2) \textbf{Stable Dreamfusion (P):} Stable Dreamfusion model \textbf{P}re-trained on a semantically close scene and finetuned to the target scene. The motivation in using Stable Dreamfusion (P) is to have a pre-trained model as a point of comparison against \ourmethod{} model. 

\subsubsection{Qualitative Evaluation}

We show out-of-distribution generation results for 8 different scenes in Fig.~\ref{fig:ood}. The inset images in the upper left of row 1 of Fig.~\ref{fig:ood} are the scenes generated zero-shot by our method, \emph{with no optimization}, when provided with the out-of-distribution prompt. The model chooses the \textit{semantic nearest neighbour} from its training data as the initial guess for out-of-distribution prompts. For example, when asked for a ``golden blender" and ``glacier knife", our model generates a scene with ``tiger striped toaster", which is the only related kitchenware appliance in the model sees during training. We pretrain the Stable Dreamfusion(P) baselines to the same scenes predicted by our model zero-shot. The pretrained scenes for Stable Dreamfusion(P) are given as insets in the upper left of row 3 and 5 in Fig.~\ref{fig:ood}.  %  We observe that our model converges to all of the target scenes within at most 20 epochs (row 1, Fig.~\ref{fig:ood}).  

By finetuning on a small number of epochs for each out-of-distribution target scene using score distillation sampling, our method can converge much faster to the target scene than the baseline DreamFusion models. In row 2 and 3 of Fig.~\ref{fig:ood}, we see that both Dreamfusion(S) and (P), barely learn the target shape for the same amount of training budget as our method. In rows 4 and 5 of Fig.~\ref{fig:ood} we let the baselines train to convergence, and even then the quality of the converged baseline scenes are worse or at best comparable to our model's generation quality. On average we see a 5x speedup in convergence.

Importantly, DreamFusion(P) which is pre-trained to \textbf{the same zero-shot predictions of our model} is unable to be fine-tuned to the target scene as efficiently and at times get stuck in suboptimal local minima close to the initialization (see ``yarn skateboard" row 3 and 5 in Fig.~\ref{fig:ood}). This demonstrates that HyperFields learns a semantically meaningful mapping from text to NeRFs that cannot be arbitrarily achieved through neural optimization. We further explore the smoothness of this mapping through interpolation experiments in Sec.~\ref{supp:interp} of the appendix. 
%\label{sec:gen_ood}

\begin{figure}[H]
\centering
\includegraphics[width=0.48\textwidth]{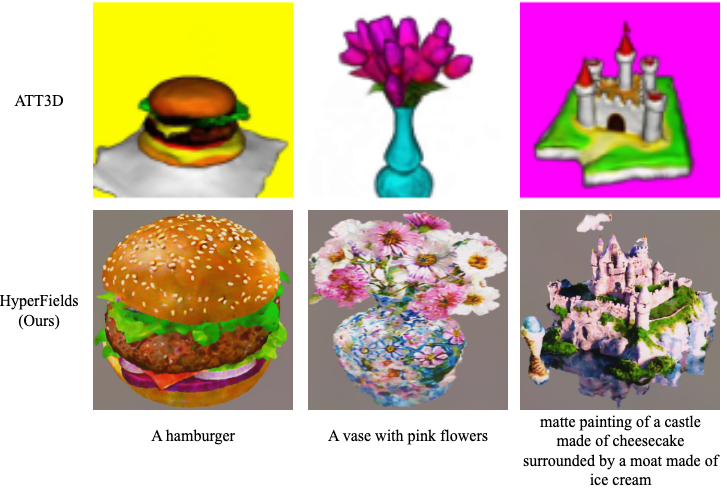}
\caption{\textbf{Visual Comparison to ATT3D.} We visually compare scenes packed into HyperFields against the same scenes shown in ATT3D. NeRF distillation allows \ourmethod{} to inherit the high generation quality of Prolific Dreamer, so the scenes we generate are of higher visual quality and complexity.} 
\label{fig:att3d_compare}
\end{figure}

\begin{figure}[h]
\centering
\includegraphics[width=0.4\textwidth]{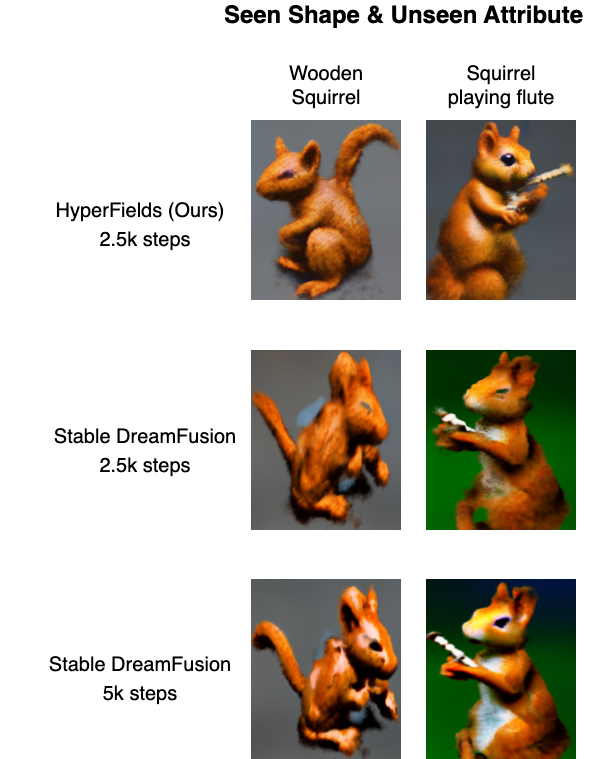}
\caption{\textbf{Prolific Dreamer OOD Comparison.} We finetune a HyperFields model trained on the scenes in Fig.~\ref{fig:complexpack} on novel attributes, and compare against the Stable DreamFusion baseline trained for the same number of steps and \textbf{double} the number of steps.} 
\label{fig:att3d_ood}
\end{figure}

\subsubsection{Quantitative Evaluation}
\label{sec:study}

\begin{table}[h]
    \centering
    \resizebox{0.5\textwidth}{!}{
    \begin{tabular}{lcccc}
        \toprule
        Model/Metric & \multicolumn{2}{c}{CLIP Top-3} & KID $\downarrow$ & SSIM $\uparrow$ \\
        \cmidrule(r){2-3}
                     & Precision & Recall & & \\
        \midrule
        Stable DreamFusion & 0.50 & 0.37 & 0.17 & 0.55 \\
        Our Model          & 0.77 & 0.63 & 0.13 & 0.62 \\
        \bottomrule
    \end{tabular}
    }
    \caption{The HyperFields-generated renders for the out-of-distribution prompts (see Fig.~\ref{fig:ood}) demonstrate superior performance compared to Stable DreamFusion's renders across multiple metrics.}
    \label{tab:performance_metrics}
\end{table}
Additionally, in order to get a quantitative evaluation of our generation quality for the out-of-distribution prompts (in Fig. \ref{fig:ood}) we conduct a human study where we ask participants to rank the render that best adheres to the given prompt in descending order (best render is ranked 1). We compare our method's generation with 33 different DreamFusion models. 1 is trained from scratch and the other 32 are finetuned from checkpoints corresponding to the prompts in section ~\ref{sec:ood_supp}. Of these 33 models we pick the best model for each of the out-of-distribution prompts, so the computational budget to find the best baseline for a given prompt is almost 33x that our of method. Note each of these models, including ours, are trained for the same number of steps. We report average user-reported rank for our method and the average best baseline chosen \textit{for each prompt} in Tab.~\ref{tab:study}. We outrank the best DreamFusion baseline consistently across all our out-of-distribution prompts.  

Furthermore, in Tab.~\ref{tab:performance_metrics} we evaluate our method's out-of-distribution generation quality using precision and recall metrics for top-3 CLIP retrieval tasks. For each given out-of-distribution prompt (in Fig. \ref{fig:ood}), we check if the associated render is among the top three retrievals according to CLIP's ranking. Additionally, we assess the quality of our renders using KID and SSIM scores. Across all these quantitative metrics, HyperFields outperforms the stable DreamFusion baseline.

%Additionally in Sec.~\ref{sec:study} of the appendix we have a user study favourably comparing our generation to that of the baselines.

\subsection{HyperFields with Prolific Dreamer Teachers}

\label{sec:prolificdreamer}
NeRF distillation training means that our pipeline is agnostic to the choice of text-to-3D model, and thus can inherit high-quality generation properties from the latest open-source models. We demonstrate this in Fig.~\ref{fig:adddreamerfields}, where we generate teacher NeRFs using Prolific Dreamer \citep{wang2023prolificdreamer} and distill them into a single HyperFields model. Our model generates the distilled scenes with virtually no quality degradation. We provide a visual comparison of our generations against the same scenes from ATT3D in Fig.~\ref{fig:att3d_compare}. 

We also show accelerated out-of-distribution convergence of our high-quality HyperFields model in Fig~\ref{fig:att3d_ood}. Note that even with double the amount of training, the Stable DreamFusion baseline is unable to match our model's generation quality. 

%We compare to the quality of scenes generated by ATT3D and \ourmethod{} with 
%\ourmethod{} can learn high-quality, complex scenes with no loss of quality and similar generalization abilities. Following the training pipeline in Fig.~\ref{fig:overview}, in stage 1 we train individual Prolific Dreamer models . In stage 2 we distill the Prolific Dreamer models into HyperFields. We show a subset of the scenes learned by a single \ourmethod{} model in Fig.~\ref{fig:adddreamerfields}.

%In addition to the scenes in Fig.~\ref{fig:dreamerfields}, we show 30 more scenes learned by a single HyperFields model in Fig.~\ref{fig:adddreamerfields}. We train a single HyperFields model to learn in total 50+ scenes (we show 38 of those scenes Fig. ~\ref{fig:dreamerfields}, Fig. ~\ref{fig:adddreamerfields})
%\begin{figure}
%\centering
%\includegraphics[width=\textwidth]{images/dreamer_pack.drawio.png}
%\caption{\textbf{Distilling Prolific Dreamer into HyperFields:} Demonstrates that HyperFields can be made to learn high quality complex scenes simply by using better teacher model like Prolific Dreamer.} 
%\label{fig:dreamerfields}
%\end{figure}

% \begin{center}
%     \centering
%     \includegraphics[width=0.7\textwidth]{images/additional_packing_dreamer.drawio.jpg}
%     \captionof{figure}{\textbf{Prolific Dreamer scenes distilled into HyperFields:} depicting various complex poses of animals, thereby underscoring the ability of a single HyperFields model to learn multiple complex scenes. } 
%     \label{fig:adddreamerfields}
% \end{center}%

\subsection{Amortization Benefits} 

\label{sec:amortization}
The cost of pre-training \ourmethod{} and individual teacher NeRFs is easily amortized in both in-distribution and out-of-distribution prompts. Training the teacher NeRFs is not an additional overhead; it's the cost of training a DreamFusion model on each of those prompts. The only overhead is the NeRF distillation training in stage 2 (Fig.~\ref{fig:overview}), which takes roughly two hours. This overhead is offset by our ability to generate unseen combinations in a feedforward manner.

For comparison, the DreamFusion baseline takes approximately 30 minutes to generate each test scene in Fig.~\ref{fig:colormatrix}, totaling $\sim$14 hours for all 27 test scenes. Our model can generate all 27 test scenes in less than a minute, making it an order of magnitude faster than DreamFusion, even with the 2 hour distillation overhead.

Our method's ability to converge faster to new out-of-distribution prompts leads to linear time-saving for each new prompt. This implies a practical use case of our model for rapid out-of-distribution scene generation in a real world setting. As shown in Fig.~\ref{fig:ood}, the baseline's quality only begins to match ours after 3-5x the amount of training time.

\begin{figure}[h]
    \centering
    \includegraphics[width=0.45\textwidth]{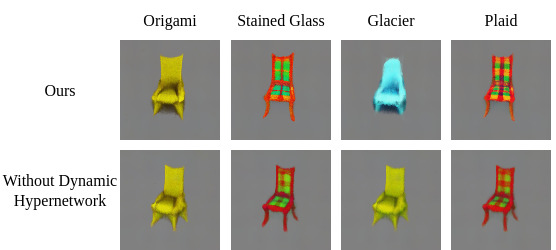}
    \caption{\textbf{Dynamic Hypernet Packing.} Without dynamic conditioning, the network collapses the origami/glacier attributes and stained glass/plaid attributes.}
     \label{abl:packing}
\end{figure}

\begin{figure}[h]
    \centering
    \includegraphics[width=0.35\textwidth]{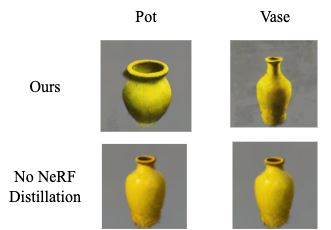}
    \caption{\textbf{NeRF Distillation.} We compare packing results when training with score distillation (``No NeRF Distillation") versus our NeRF distillation method (``Ours"). The iterative optimization of score distillation causes mode collapse in geometry.\vspace{-0.5cm}}
    \label{abl:distillation}
\end{figure}

\subsection{Ablations}
\label{sec:ablations}

% We demonstrate the necessity of NeRF distillation and our dynamic hypernetwork architecture through ablations, shown in Fig.~\ref{abl:distillation} and \ref{abl:packing} and Sec. \ref{supp:ablations} in the appendix.

We ablate on the activation conditioning in our dynamic hypernetwork (``without dynamic hypernetwork'') in Fig.~\ref{abl:packing}. Row 2 shows that even in the simple case of 4 scenes the static hypernetwork collapses the ``glacier" and ``origami" styles, and the ``plaid" and ``stained glass" styles.

If we attempt to pack the dynamic hypernetwork using just Score Distillation Sampling (SDS) from DreamFusion, we experience a type of mode collapse in which the SDS optimization guides similar shapes towards the same common geometry. See Fig.~\ref{abl:distillation} for an example of this mode collapse.

%\subsection{Limitations}
%\label{sec:limitations}
%There are a few key limitations of HyperFields. First, the quality of the  scenes used in training  and consequently the quality of the generalization scenes is constrained by the quality of the teacher NeRFs. Any systematic artifacts in the teacher NeRFs will be learned by the hypernetwork. Second, the dynamic hypernetwork while allows for flexibility in representation of various scenes consumes a large amount of GPU RAM making our model trainable  only on 48GB or higher GPU RAMs. Finally, our model cannot generate scenes zero-shot without any optimization in an open vocabulary setting, an open problem for future research.

%% file: conclusion.tex
\section{Conclusion}
We present \ourmethod{}, a novel framework for generalized text-to-NeRF synthesis, which can produce individual NeRF networks in a single feedforward pass. Our results highlight a promising step in learning a general representation of semantic scenes.  Our novel dynamic hypernetwork architecture coupled with NeRF distillation learns an efficient mapping of text token inputs into a smooth and semantically meaningful NeRF latent space. Our experiments show that with this architecture we are able to fit over 100 different scenes in one model, and predict high quality unseen NeRFs either zero-shot or with a few finetuning steps. Comparing to existing work, our ability to train on multiple scenes greatly accelerates convergence of novel scenes.  In future work we would like to explore the possibility of generalizing the training and architecture to achieving zero-shot open vocabulary synthesis of NeRFs and other implicit 3D representations. %\new{We hypothesize that such a design choice equips the model with added flexibility to escape undesirable local minima.}

\section{Limitations}
Our model is trained through distillation from teacher models, thus, the quality of our generated scenes is bound by the quality of the current state-of-the-art open source models. Similarly, our model inherits the limitations of these SOTA models. For instance, it is well known that Stable Diffusion struggles with long prompts with complex compositionality and janusing, which are also limitations of our model.

\section*{Impact Statement}
As mentioned above our model inherits limitations from the teacher models. Similarly, our model inherits  potential harmful biases of the teacher models. Any stereotypes or biases from the teacher models will be reproduced by our model.

\section*{Acknowledgements}
This work was partially supported by NSF grants CNS-1956180 and 2304481, and BSF grant 2022363. Additional support was provided by gifts from Snap Research, Adobe Research, and Google Research. We also thank Adam Bohlander for his assistance in setting up the computing infrastructure.

%% file: supp_paper.tex
%\begin{appendices}
\appendix
\onecolumn 

\section{Model Details}

\textbf{Baselines:} Our baseline is  6 layer MLP with skip connections every two layers. The hidden dimension is 64. We use an open-source \href{https://github.com/ashawkey/stable-dreamfusion}{re-implementation} \cite{stable-dreamfusion} of DreamFusion as both our baseline model and architecture predicted by \ourmethod{}, because the original DreamFusion works relies on Google's Imagen model which is not open-source. Unlike the original DreamFusion, the re-implementation uses Stable Diffusion (instead of Imagen). We use Adam with a learning rate of 1e-4, with an epoch defined by 100 gradient descent steps.

\textbf{HyperFields:} The architecture is as described in Figure 2 in the main paper. The dynamic hypernetwork generates weights for a 6 layer MLP of hidden dimension 64. The transformer portion of the hypernetwork has 6 self-attention blocks, each with 12 heads with a head dimension of 16. We condition our model with BERT tokens, though we experiment with T5 and CLIP embeddings as well with similar but marginally worse success. Similar to the baseline we use Stable Diffusion for guidance, and optimize our model using Adam  with the a learning rate of 1e-4. We will release open-source code of our project in a future revision of the paper. 

We use the multiresolution hash grid developed in InstantNGP \citet{instantngp} for its fast inference with low memory overhead, and sinusoidal encodings $\gamma$ to combat the known spectral bias of neural networks \citep{spectralbias}. The NeRF MLP has 6 layers (with weights predicted by the dynamic hypernetwork), with skip connections every two layers. The dynamic hypernetwork MLP modules are two-layer MLPs with ReLU non-linearities and the Transformer module has 6 self-attention layers. Furthermore, we perform adaptive instance normalization before passing the activations into the MLP modules of the dynamic hypernetwork and also put a stop gradient operator on the activations being passed into the MLP modules (as in figure \ref{fig:high_level}).

\section{Packing}

\begin{figure*}[h]
    \centering
    
    \includegraphics[width=0.8\textwidth]{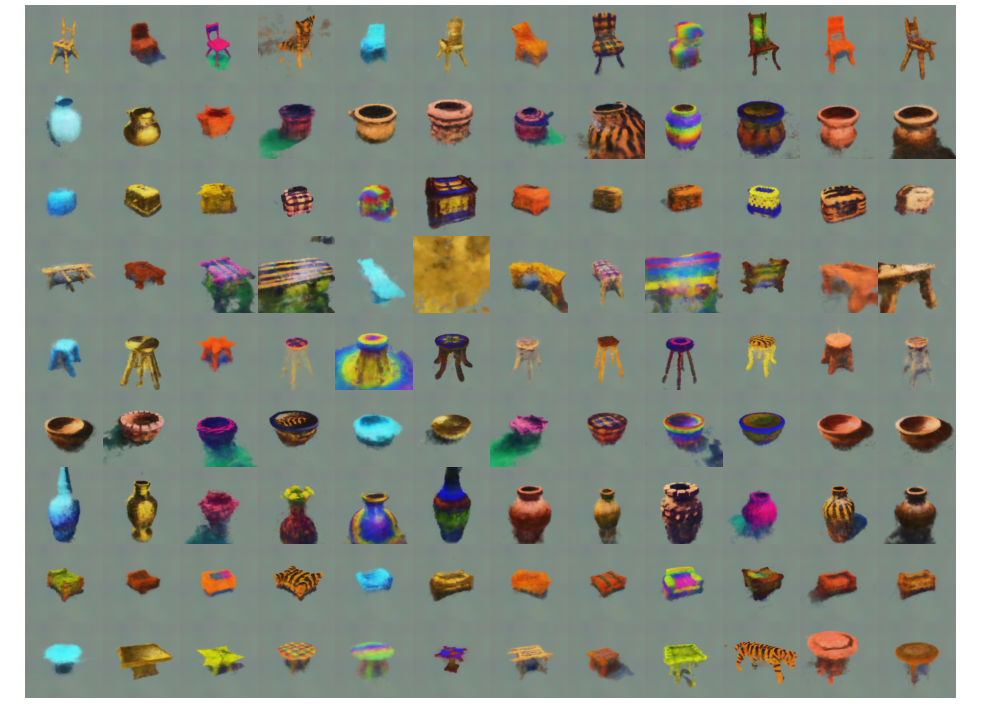}
    \caption{\textbf{ Prompt Packing.} Our dynamic hypernetwork is able to pack 9 different objects across 12 different prompts for a total of 108 scenes. Dynamic hypetnetwork coupled with NeRF distillation enables packing these scenes into one network.}
    \label{fig:complexpack}
\end{figure*}

\section{In-Distribution Generalization with Complex Prompts}

\begin{figure*}
\centering
\includegraphics[width=0.8\textwidth]{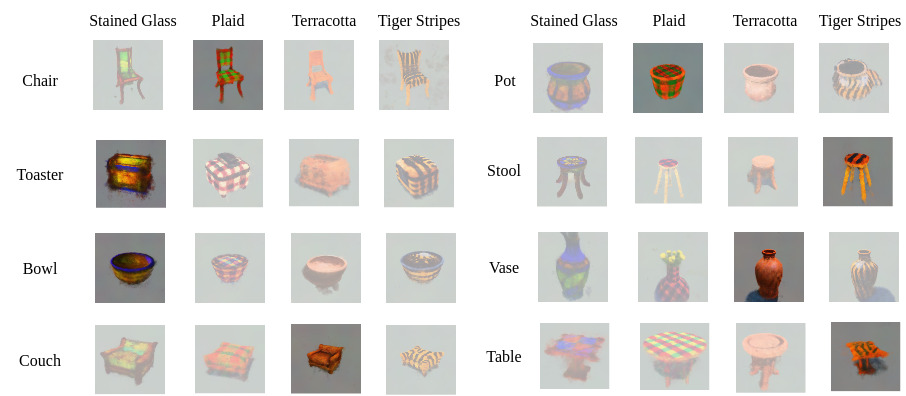}
\caption{\textbf{ Fine-Tuning In-Distribution: seen shape, seen attribute, unseen combination.} During training, the model observes every shape and color, but some combinations of shape and attribute remain unseen.
During inference, the model generalizes by generating scenes that match prompts with previously unseen combinations of shape and attribute, with small amount of finetuning (atmost 1k steps).}
\label{fig:stylematrix}
\end{figure*}

For additional attributes (``plaid", ``Terracotta" etc.), our model produces reasonable zero-shot predictions, and after fewer than 1000 steps of finetuning with SDS is able to produce unseen scenes of high quality. We show these results in Fig.~\ref{fig:stylematrix} with 8 objects and 4 styles, where 8 shape-style combinational scenes are masked out during training (opaque scenes in Fig.~\ref{fig:stylematrix}).

\begin{figure}
\centering
\includegraphics[width=0.8\columnwidth]{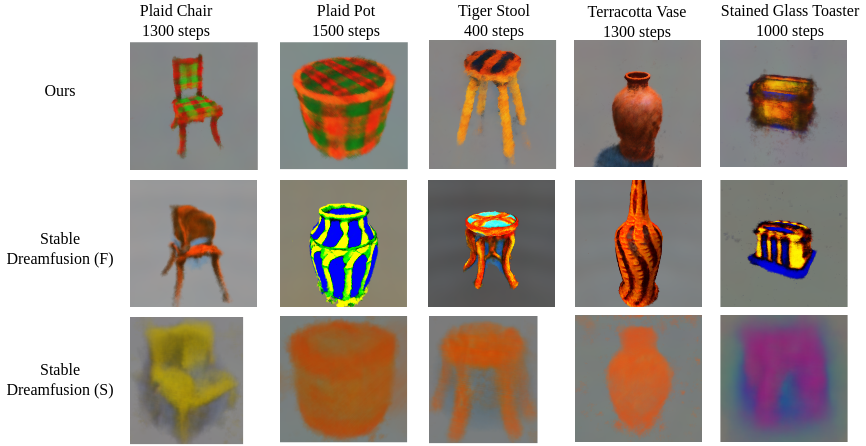}
\caption{\textbf{Generalization Comparison.} We train a single \ourmethod{} model and compare Stable DreamFusion. ``Stable DreamFusion (F)" indicates finetuning from an initialized DreamFusion model. ``Stable DreamFusion (S)" indicates the DreamFusion model trained from scratch. Zero-shot results and initializations are shown in the upper left of ``Ours" and ``Stable DreamFusion (F)", respectively. Above each column indicates the number of training epochs for each method add figures in the upper left. }
\label{fig:stylebaseline}
\end{figure}
\section{Out-of-Distribution Convergence}
\label{sec:ood_supp}
In Fig~\ref{fig:ood} we show the inference time prompts and the corresponding results. Here we provide the list of prompts used to train the model: ``Stained glass chair",  ``Terracotta chair", ``Tiger stripes chair", ``Plaid toaster", ``Terracotta toaster", ``Tiger stripes toaster", ``Plaid bowl", ``Terracotta bowl", ``Tiger stripes bowl", ``Stained glass couch", ``Plaid couch", ``Tiger stripes couch", ``Stained glass pot", ``Terracotta pot", ``Tiger stripes pot", ``Stained glass vase", ``Plaid vase", ``Tiger stripes vase", ``Stained glass table", ``Plaid table", ``Terracotta table".

Since the training prompts dont contain shapes such as ``Blender", ``Knife", ``Skateboard",  ``Shoes", ``Strawberry", ``Lamppost", ``Teapot" and attributes such as ``Gold", ``Glacier", ``Yarn", ``Victorian", we term the prompts used in Fig~\ref{fig:ood} as out-of-distribution prompts--as the model does not see these shapes and attributes during training.

\section{User Study Renders}

We link to the images (including baselines) used in the user study described in \cref{sec:study} \href{https://drive.google.com/file/d/1CTbGXMLVnqTsupslpD6XK0fuOlmwynbv/view?usp=sharing}{here}. All renders are taken from the same camera angle and the baseline scenes are finetuned with the same number of iterations as our model.

% \begin{figure*}
%     \centering
%     \includegraphics[height=\textheight]{images/studyimages.png}
%     \caption{\textbf{User Study Images.} We link to the images (including baselines) used in the user study described in \cref{sec:study}. All renders are taken from the same camera angle and the baseline scenes are finetuned with the same number of iterations as our model.}
%     \label{fig:studyimages}
% \end{figure*}

% \section{Additional Ablation on Dynamic Hypernetwork}
% We train both the model used in figure \ref{fig:add_aba} simultaneously on prompts `Wooden Table' and `Origami Chair', we see that our model easily learns to generate both these scenes, while the model without the dynamic hypernetwork suffers from lack flexibility and consequently make the geometries of the `Wooden table' and `Origami Chair' similar. 
% \label{supp:interp}
% \begin{figure*}
%     \centering
%     \includegraphics[width=0.4\textwidth]{images/additional_abalation.png}
%     \caption{ \textbf{Ablation over dynamic hypernetwork:} We see that our dynamic hypernetwork can generate origami chair and wooden table, while when trained without our dynamic hypernetwork the model suffers from mode collapse and makes the origami chair have a geometry similar to the wooden table.}
%     \label{fig:add_aba}
% \end{figure*}

\section{Algorithm for training HyperFields}
\begin{minipage}{.99\linewidth} % Adjust the width to your liking
\vspace{-1ex}
\rule{\linewidth}{1pt}
  %\captionof{algorithm}{Training HyperFields with NeRF Distillation}
  \label{alg:training}
  \vspace{-2ex}
  \rule{\linewidth}{1pt}
  \begin{algorithmic}[1]
    \REQUIRE $\mathcal{T}$ = $\{\mathcal{T}_{1}, \mathcal{T}_{2}, \cdots \mathcal{T}_{N} \}$ \hfill $\vartriangleright$ \scriptsize{Set of text prompts}
    \REQUIRE $\mathcal{C}$ \hfill $\vartriangleright$ \scriptsize{Set of Camera view points}
    \REQUIRE $\theta_{1}, \theta_{2}, \cdots \theta_{N}$ \hfill $\vartriangleright$ \scriptsize{pre-trained NeRFs  }
    \REQUIRE $\phi_{HF}$  \hfill $\vartriangleright$ \scriptsize{Randomly initialized HyperFields }
    \REQUIRE $\mathcal{R}$  \hfill $\vartriangleright$ \scriptsize{Differentiable renderer function }
    \FOR{each step}
    \STATE $\mathcal{T}_{l}$, $\mathcal{T}_{m}$,$\mathcal{T}_{n} \sim \mathcal{T}$ \hfill $\vartriangleright$ \scriptsize{Sample text prompts from $\mathcal{T}$  }
      \FOR{$\mathcal{T}_{i} \in \{\mathcal{T}_{l}$, $\mathcal{T}_{m}$,$\mathcal{T}_{n}\}$}
        \STATE $\mathcal{C}_{i} \sim  \mathcal{C}$
        \STATE ${\mathcal{I}_{i}} = \mathcal{R}(\theta_{i}(\mathcal{C}_{i})) $ \hfill $\vartriangleright$ \scriptsize{ $i^{th}$ nerf renders image for given camera $\mathcal{C}_{i}$  }
        \STATE ${\mathcal{I}_{i}}^{'} = \mathcal{R}(\phi_{HF}(\mathcal{T}_{i}, \mathcal{C}_{i})) $ \hfill $\vartriangleright$ \scriptsize{ Condition $\phi_{HF}$ on $i^{th}$ prompt}
        \STATE $\mathcal{L}_{i}$ = $({\mathcal{I}_{i}} - {\mathcal{I}_{i}}^{'})^{2} $ 
      \ENDFOR
      \STATE $\mathcal{L}_d = \sum \limits_{i \in \{l,m,n\}}  \mathcal{L}_{i}$
    \ENDFOR
  \end{algorithmic}
\end{minipage}
% \section{Additional Out-of-Distribution Results}
% To add more variability to our  OOD (out-of-distribution) results from Fig. 5 we add animal prompts. We consider ‘wooden squirrel’ and ‘squirrel playing flute’, for both these prompts the geometry generated by HyperFields seems to be better than the StableDreamFusion outputs in our opinion. The corresponding renders are in Fig ~\ref{fig:add_ood}. These prompts are OOD because the model has not seen the attribute wooden and playing flute in its training set. We optimize our HyperFields model using SDS loss.

% \begin{figure*}
%     \centering
%     \includegraphics[width=0.4\textwidth]{images/fancy_ood.jpg}
%     \caption{ \textbf{Comparison to Stable DreamFusion.} We visually compare to the Stable DreamFusion baseline model trained from scratch using SDS on the prompts ``wooden squirrel" and ``squirrel playing the flute". Through NeRF distillation, HyperFields reconstructs the ProlificDreamer models with high fidelity, and thus achieves better scene quality than SDS-based optimization.}
%     \label{fig:add_ood}
% \end{figure*}

\section{HyperFields Trained on Additional Prolific Dreamer Teachers }
\begin{figure*}
     \centering
     \includegraphics[width=1\textwidth]{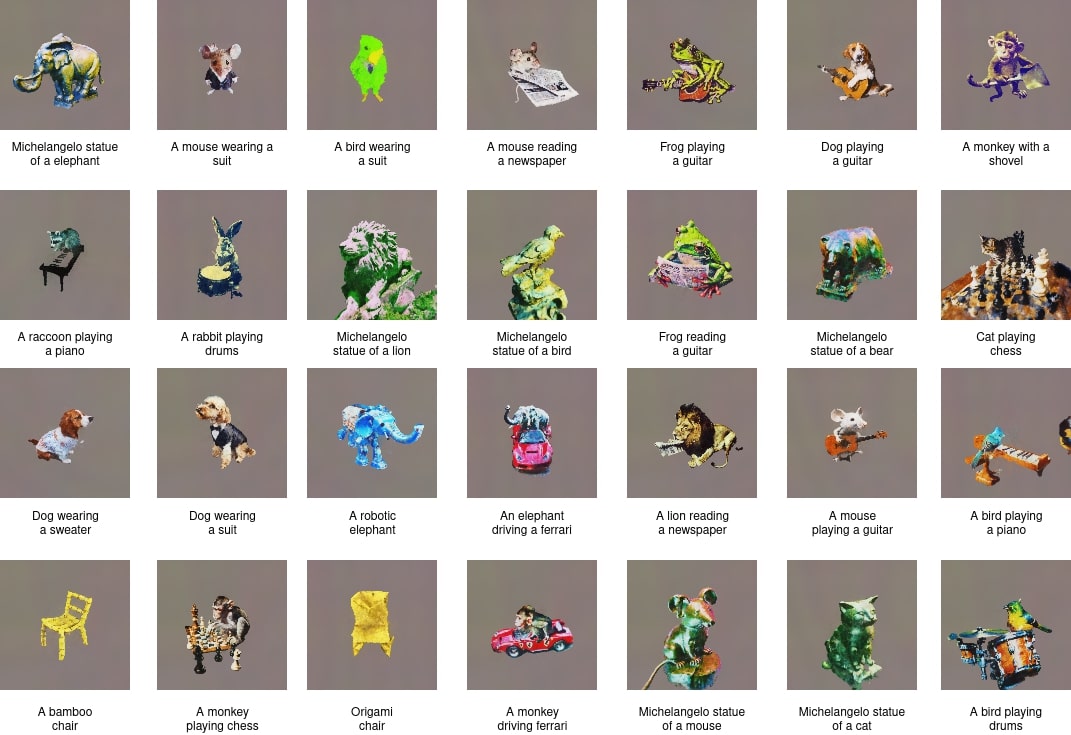}
     \caption{Additional set of Prolific Dreamer scenes distilled into HyperFields model, showcasing the ability of Hyperfields to learn a significantly diverse set of geometries.}
     \label{fig:new_dreamer}
 \end{figure*}

In addition to the scenes shown in Fig. ~\ref{fig:adddreamerfields}, train HyperFields on a different set of ProlificDreamer teachers and the scenes generated by our single HyperFields model is shown in Fig. \ref{fig:new_dreamer}. This demonstrates the ability of HyperFields to learn another varied set of scenes with complex geometries.

\section{Multi-View Consistency of Generated Scenes }
\begin{figure*}
     \centering
     \includegraphics[width=1\textwidth]{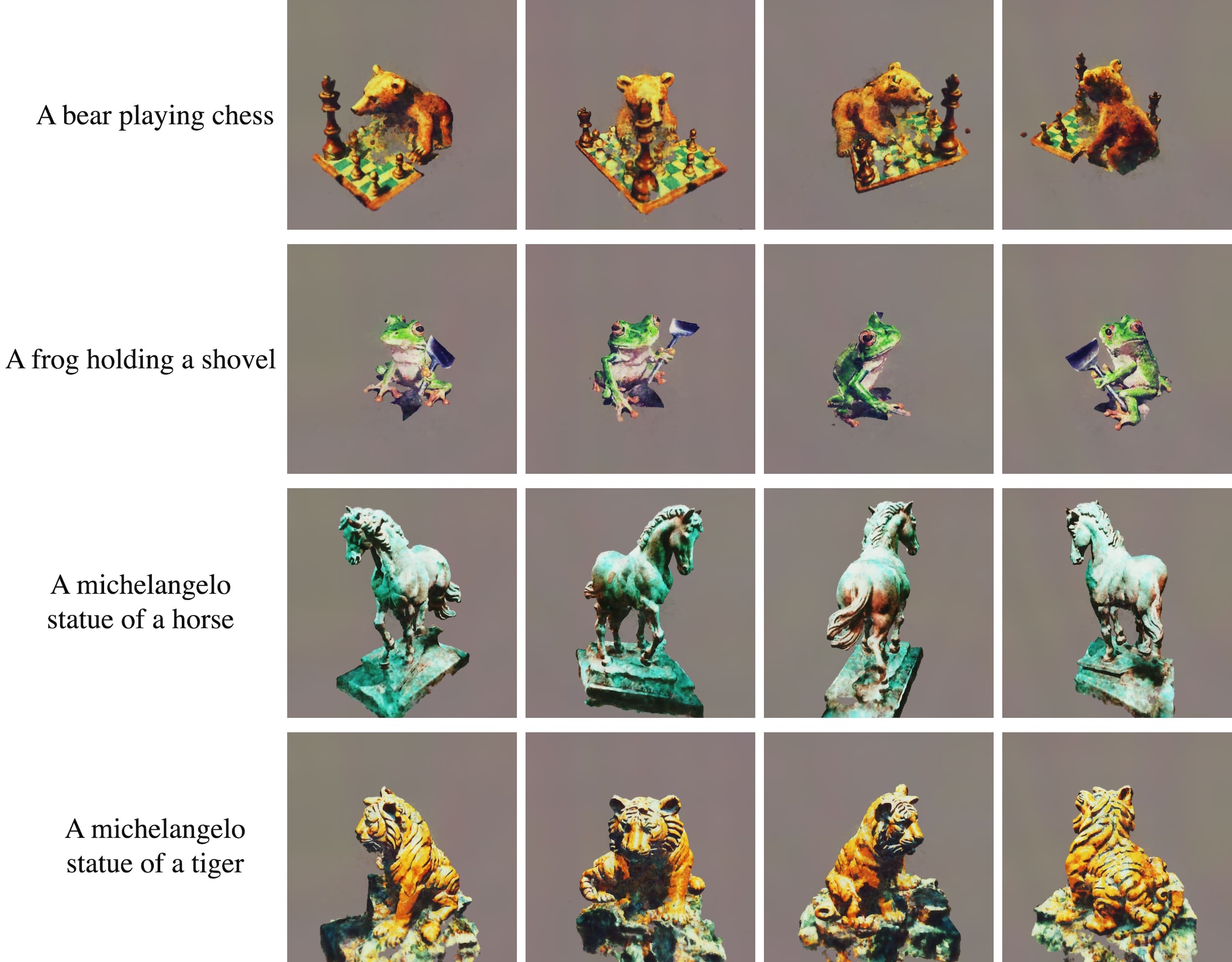}
     \caption{Renders of various scenes generated by HyperFields from various camera poses. The geometry is well formed and consistent across multiple views.}
     \label{fig:multiview}
 \end{figure*}

In Fig. ~\ref{fig:multiview}, we show multiple scenes generated  by a single HyperFields model from various camera poses. Across multiple views we see that geometry is well formed and consistent with the geometry in other views. 

\section{BERT Token Interpolation}
\label{supp:interp}
\begin{figure*}
    \centering
    \includegraphics[width=\textwidth]{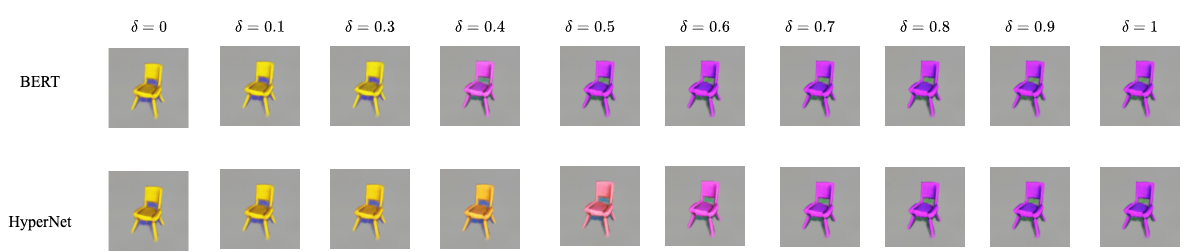}
    \caption{\textbf{BERT Token Interpolation.} We show results of interpolating the BERT tokens corresponding to the prompts ``yellow chair" and ``purple chair". In contrast, interpolation on the level of the hypernetwork (``HyperNet") is smoother than interpolating the BERT tokens. }
    \label{fig:bertinterpolation}
\end{figure*}

Another option for interpolation is to interpolate the input BERT embeddings fed in the our Dynamic HyperNet. We show results in \cref{fig:bertinterpolation} where we interpolate across two chair colors in a Dynamic HyperNet trained on only chair colors. The interpolation is highly non-smooth, with a single intermediate color shown at $\delta=0.4$ and discontinuities on either end at $\delta-0.3$ and $\delta=0.5$. On the other hand, our HyperNet token interpolation shown in Figure 10 demonstrates a smooth and gradual transition of colors across the interpolation range. This demonstrates that our HyperNet learns a smoother latent space of NeRFs than the original BERT tokens correspond to.